# Leveraging human Domain Knowledge to model an empirical Reward function for a Reinforcement Learning problem

Dattaraj Jagdish Rao
Persistent Systems Ltd.
dattaraj_rao@persistent.co.in

16-SEPT-2019

## ABSTRACT

Traditional Reinforcement Learning (RL) problems depend on an exhaustive simulation environment that models real-world physics of the problem and trains the RL agent by observing this environment. In this paper, we present a novel approach to creating an environment by modeling the reward function based on empirical rules extracted from human domain knowledge of the system under study. Using this empirical rewards function, we will build an environment and train the agent. We will first create an environment that emulates the effect of setting cabin temperature through thermostat. This is typically done in RL problems by creating an exhaustive model of the system with detailed thermodynamic study. Instead, we propose an empirical approach to model the reward function based on human domain knowledge. We will document some rules of thumb that we usually exercise as humans while setting thermostat temperature and try and model these into our reward function. **This modeling of empirical human domain rules into a reward function for RL is the unique aspect of this paper.** This is a continuous action space problem and using deep deterministic policy gradient (DDPG) method, we will solve for maximizing the reward function. We will create a policy network that predicts optimal temperature setpoint given external temperature and humidity.

*All the code used in this paper is available at the link below on Google Collaboratory as a Python notebook.*
*https://colab.research.google.com/drive/1JQxLS2DSv5mO1i_68pd_uOQVNwCy1kEw*

***Keywords***: reinforcement learning, reward function, domain knowledge, temperature control, ship, air conditioning

## THE PROBLEM STATEMENT

More than 30% of power generated on a cruise ship is consumed by the heating, ventilation and air conditioning (HVAC) system. Commercial and Navy ships have to run hundreds of miles and conserve fuel till the next port for refill. Any fuel saving especially on comfort items like air conditioning can be very useful in reducing operating costs. However, due to high humidity at sea, appropriate temperature setting is essential so that the air quality does not become unhealthy. This is a unique optimization problem where we want to minimize fuel consumption while maintaining the temperature and humidity to maintain human level of comfort.

Typically, thermostat in the cabin of ship is set to a fixed temperature like 22 deg C. Irrespective of the outside temperature and humidity the air conditioning (AC) system has to work towards making the temperature to this limit. The AC system has to blow cold air to reduce the room temperature when outside air temperature is high and also has to remove moisture to lower the relative humidity in the room air to get it within comfort limits [3][4]. Many times, when the temperature is high and humidity high – it may be ok to set the temperature at a higher limit to save on power consumption.

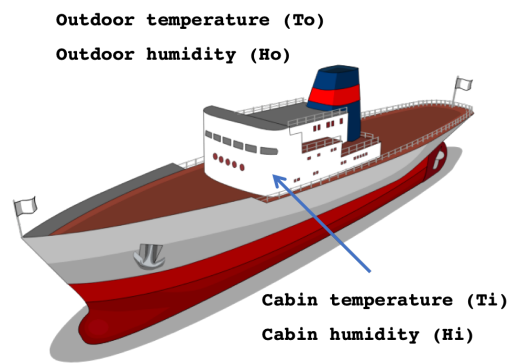

**Figure 1 Schematic of the problem**

We will build a RL agent that considers these factors and tries to provide an optimal temperature setting given a fixed setpoint. In this paper, the focus will be on the RL agent and the algorithm for training the agent. We are working with a major shipyard to build a prototype for this automated temperature correction model into their ship AC controllers. This system-level solution is out of scope of this paper.

For this study, we will keep the problem very simple and focus only on ships in Asian countries where the outdoor temperatures are high, and the main focus is on cooling the inside cabin rather than heating. Based on above schematic, the state is represented by outdoor temperature and humidity. These are the factors we will use to understand how to set inner cabin temperature. The parameter that we control is the cabin temperature (Ti). For this case, we will assume the relative humidity inside cabin needs to be maintained at a constant value of 30% as recommended by ASHRAE [1].

In a real production implementation, we will consider many additional factors like air quality index, power consumption, wind speed, etc. However, for now we will consider the above simplified problem and use RL to model it.

## REINFORCEMENT LEARNING

Reinforcement Learning is all about training a ML agent that learns from experience by directly sampling against the real world or a simulated environment. The RL problem can be formulated as a Markov Decision Process (MDP) as shown in figure below.

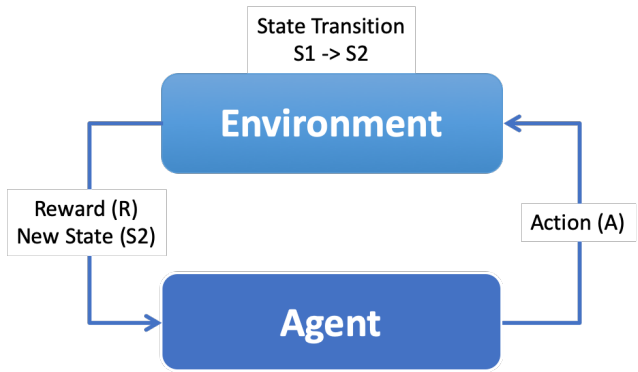

Figure 2 Basics of Reinforcement Learning

The agent interacts with the environment and takes an action (A) on the environment to cause its state to transition from state S1 to S2. After going through many such cycles the agent starts to learn the pattern of actions that maximize the rewards. [6]

There are 2 approaches to solving RL problem:

1- <u>Policy gradients</u>: Here we build a model (neural network) that maps the state space to action space. For the state defined by discrete or continuous variables – the policy network will give probability for each action. The network is trained using backpropagation and gradient ascent to maximize the value function. The value function represents the total discounted future rewards for that state. After being trained the network will now be able to predict actions for each state. Policy gradients work best on continuous and stochastic action spaces.

2- <u>Value iteration</u>: Here we run multiple iterations and collect the quality value (Q-Value) for each state-action combination. The most popular example of this method is Q-Learning. Here we store the Q-values for different states and use a technique like dynamic programming to select action based on highest Q-value. This method is further extended by a deep learning driven Deep Q-Network (DQN) approach which uses a neural network to estimate Q-value for a state-action pair. DQN has shown amazing results particularly in the gaming space with agents defeating human experts in games like Chess and Go [6].

Value iteration method are popular, but they work best when the action space is discrete – as is case with gaming (actions are moving pawns on a discrete board). For our temperature control problem, it's a continuous action space with the output variable having a continuous value.

For this particular problem we will use a popular method known as Deep Deterministic Policy Gradients (DDPG) [2]. It is a popular paradigm known as actor-critic architecture. Policy gradients suffer from the fact that they use the same policy that is being learned to collect more training data. Hence, it's very possible to get a bad policy and keep collecting bad training data and go off track. Actor critic methods try and design around this problem using 2 complimenting neural networks. The actor network actually learns the policy to predict action. The critic network takes the state and predicted action and estimates the Q-value. Thus, this method tries to understand the action as well as how good that action is to a state.

## MODELING THE ENVIRONMENT

For our problem we will first model an environment with empirical rules on adjusting states and providing rewards for state transitions. State is represented by the outdoor temperature and humidity that we will assume will be sensed by sensors that are part of our overall system solution. The action our agent will take is to suggest an indoor room temperature. This will be the optimal setpoint we want our thermostat in ship cabin to be set automatically to. We will also provide our usual fixed temperature setpoint, so that the setting will not go beyond specific limits of this value. This is just a safeguard we will use to make sure our system does not overshoot from the desired value.

Key piece of an environment model is the reward function. Typically, in RL problems, the environment simulation captures the physics of the problem domain. Here instead of trying to model the system thermodynamics we take an empirical approach. Our main purpose in building the environment is to figure out by changing temperature to certain levels – how will it affect the power consumption by AC and human comfort levels. We will use human domain knowledge to try and identify these "rules" and use these to define our reward function.

**USING DOMAIN TO MODEL REWARD FUNCTION**

First, we will model our reward function such that any violations of limits are penalized by negative reward. RL learns from trail and error, hence we cannot directly set upper and lower bounds for values the agent will sample. These bounds need to be set through the reward function. We define a band of temperatures in which our setpoint should be – let's say this is 2 deg C above and below the fixed setpoint we normally set the cabin temperature at – which we will assume to be 22 deg C. Next, the reward function should reflect power spent by the AC, so that if we spend less effort, we get bigger reward. We want to minimize the temperature difference between outside and inside so that less effort is spent by the AC system. Also, the effort spent by AC is inversely proportional to outdoor humidity. More humid outside the more effort your AC has to spend to remove moisture from air and bring to comfortable levels. Let's capture some of these rules.

*RULES from domain knowledge*

- Fixed setpoint = 22 deg C
- Low limit for RL setpoint = 20 deg C
- Upper limit for RL setpoint = 24 deg C
- Larger the temperature difference between room and outdoor, larger is the power consumed by AC
- Larger the outdoor humidity, larger is AC power consumed
-

We will model our reward function in the environment following these domain rules and train our RL agent based on this reward function. After training, we will run a simulation with different variations in outdoor temperature and humidity and compare fixed setpoint vs agent-controlled setpoint.

**SIMULATION RESULTS**

We ran a simulation with above environment with several variations in outdoor temperature and humidity. We used this environment to train a RL agent with the DDPG algorithm using a fixed setpoint of 22 deg C. Below are results for a simulation with different variations in outdoor temperature and humidity. Each point on the chart represents a day with particular To and Ho and comparison between fixed setpoint (blue line) and agent-controlled temperature (red line). After the simulation with agent, we tried to estimate the power consumption by AC in each of the 2 cases. We did not get into the exact thermodynamic formula since we just want to validate the relative power consumption. We can safely assume that power consumption by AC is inversely proportional to temperature difference between outdoor and indoor. Using this factor, we find the relative change in power consumption between fixed and agent controller AC temperatures.

Below chart shows the results from this simulation. Yellow dotted line shows outdoor temperature variation, green dotted line outdoor humidity. Blue line us the fixed temperature setting, and red line is temperature that the RL agent set.

The results are in-line with results from similar problems where RL agent was used to optimize room temperature but using a detailed physics-based model. [3] [4] [5]

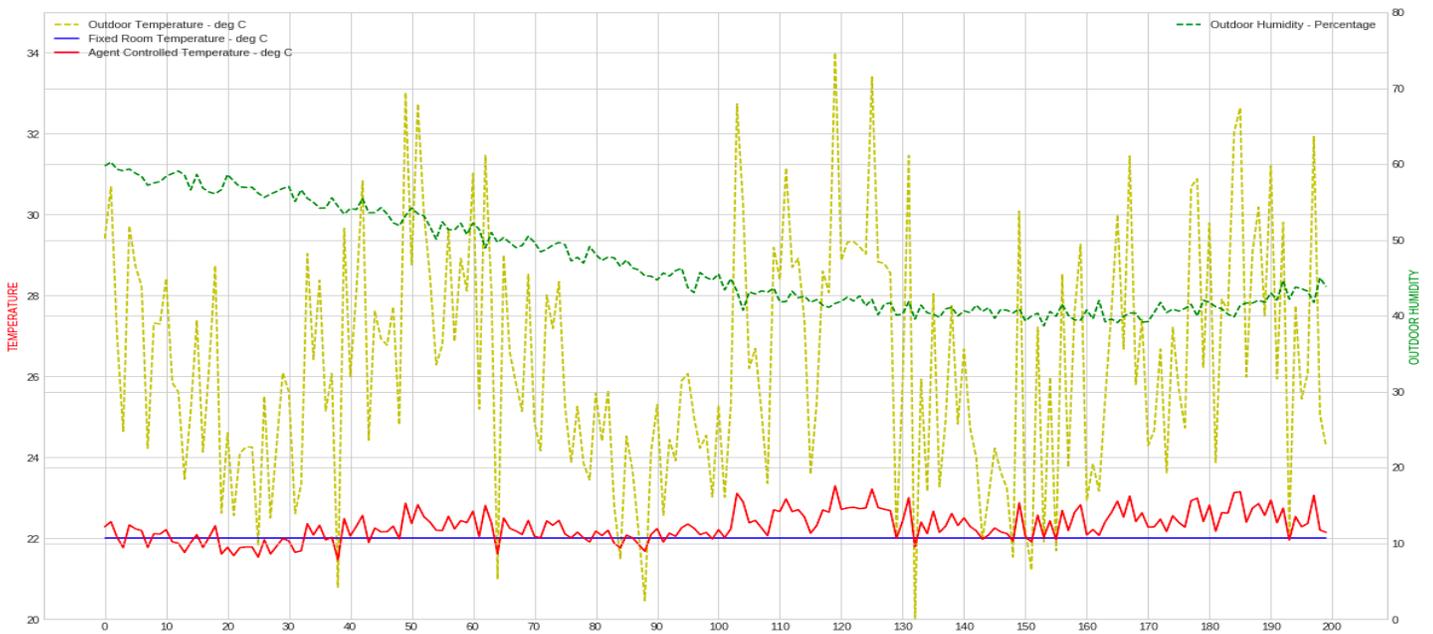

Figure 3 Results of simulation - Fixed (blue line) vs Agent-controlled temperature (red line)

## CONCLUSION

From the results of the simulation, we see that the red line shows variability in room AC temperature to match the outdoor humidity and temperature. We calculated the area between indoor and outdoor temperatures for both cases – fixed setpoint and agent-controlled setpoint temperature. Comparing the area differences between two temperature curves will give us an estimate of relative power consumption for each approach.

```
Area for fixed temperature = 867.5
Area for agent-controlled temperature = 808.3
Reduction in area = Improvement = 6.8%
```

We see that using the RL agent we get approximately 7% improvement in maintaining temperature difference between room and outdoor while maintaining the humidity and comfort constraints. A large cruise ship may consume around 60000 gallons of fuel per day – costing around $200,000. A saving of 7% in fuel will lead to a direct saving of $14,000 per day or almost a $5MM savings annually!

These results appear in-line very much with similar results when a detailed thermodynamics reward function was used. Hence, we see that a basic empirical reward function can give us very good results in training RL agent.

This basic simulation shows how using RL we can control temperature setpoint and show major improvements in power consumption by the AC system. Next step is to collect real data from a ship AC unit and evaluate if we can confirm the same savings in power consumption. Also, we will incorporate this RL agent into a real system that can control the temperature of ship AC and measure the actual power savings over time.